\documentclass[sigconf]{acmart}
\usepackage{multirow} 
\usepackage{balance}
\copyrightyear{2026}
\acmYear{2026}
\setcopyright{cc}
\setcctype{by-nc-nd}
\acmConference[CIKM '26]{Proceedings of the 35th ACM International Conference on Information and Knowledge Management}{November 7--11, 2026}{Rome, Italy.}
\acmBooktitle{Proceedings of the 35th ACM International Conference on Information and Knowledge Management (CIKM '26), November 7--11, 2026, Rome, Italy}
\acmISBN{979-8-4007-2539-5/2026/11}
\acmDOI{10.1145/3799682.3840194}  

\begin{document}

\title{A Benchmark \& Dataset for Detecting AI-Manipulated Visual Evidence in the Court System}

\author{Kelly McConvey}
\authornote{Corresponding author. Email: \texttt{kelly.mcconvey@mail.utoronto.ca}}
\affiliation{%
  \institution{University of Toronto}
  \city{Toronto}
  \country{Canada}}

\author{Sajad Ebrahimi}
\affiliation{%
  \institution{University of Toronto}
  \city{Toronto}
  \country{Canada}}

\author{Nima Jamali}
\affiliation{%
  \institution{University of Waterloo}
  \city{Waterloo}
  \country{Canada}}

\author{Jalehsadat Mahdavimoghaddam}
\affiliation{%
  \institution{University of Toronto}
  \city{Toronto}
  \country{Canada}}

\author{Matina Mahdizadeh Sani}
\affiliation{%
  \institution{University of Waterloo}
  \city{Waterloo}
  \country{Canada}}

\author{Maksym Taranukhin}
\affiliation{%
  \institution{University of British Columbia}
  \city{Vancouver}
  \country{Canada}}

\author{Wentao Zhang}
\affiliation{%
  \institution{University of Waterloo}
  \city{Waterloo}
  \country{Canada}}

\author{Jacquelyn Burkell}
\affiliation{%
  \institution{University of Western Ontario}
  \city{London}
  \country{Canada}}

\author{Yuntian Deng}
\affiliation{%
  \institution{University of Waterloo}
  \city{Waterloo}
  \country{Canada}}

\author{Karen Eltis}
\affiliation{%
  \institution{University of Ottawa}
  \city{Ottawa}
  \country{Canada}}

\author{Maura R. Grossman}
\affiliation{%
  \institution{University of Waterloo}
  \city{Waterloo}
  \country{Canada}}

\author{Vered Shwartz}
\affiliation{%
  \institution{University of British Columbia}
  \city{Vancouver}
  \country{Canada}}

\author{Ebrahim Bagheri}
\affiliation{%
  \institution{University of Toronto}
  \city{Toronto}
  \country{Canada}}

\renewcommand{\shortauthors}{Kelly McConvey et al.}

\begin{abstract}
Photographic evidence is becoming increasingly vulnerable to forms of alteration and fabrication that existing legal and technical workflows are not well equipped to evaluate. Surveillance frames, dashcam stills, and phone photographs 
may be used to establish presence, sequence, causation, damage, or identity, yet contemporary generative systems allow non-experts to alter or fabricate such images through ordinary 
prompt-based interfaces.  Existing image-forensics benchmarks provide important resources for face manipulation, classical tampering, and general synthetic-image detection, but they are not organized around the forms of visual evidence submitted in courts, the localized edits that can change what an exhibit appears to prove, or the consumer-tool threat model now facing the justice system. We introduce the CIFAR Synthetic Evidence Corpus for Detecting AI-Manipulated Images, a benchmark for evidentiary image authentication in court and justice-system contexts. The corpus contains 1,505 photographic items, including 720 authentic controls and 785 manipulated or fabricated images, spanning surveillance, dashcam, and consumer-photo imagery. Manipulations are organized into scene-condition edits, localized element edits, and full fabrications produced with contemporary generative systems. Each item is released with structured metadata covering source provenance, manipulation tier, subtype, generator, prompt template, and scene attributes, enabling controlled evaluation beyond aggregate binary detection. We also establish state-of-the-art baselines with publicly available image-manipulation detectors, showing that current systems exhibit error profiles that remain problematic for evidentiary use. The dataset, prompts, metadata manifest, code, and baseline evaluation scripts are released to support research on visual evidence authentication, information integrity, and trustworthy AI for the justice system.
\end{abstract}


\maketitle

\section{Introduction}
\label{sec:Introduction}
Courts increasingly receive photographs whose authenticity can no longer be taken for granted, including surveillance frames offered to show that a person was present at a location, dashcam stills used to reconstruct traffic incidents, and phone photographs submitted to document damage, injury, possession, or the condition of a scene. In each case, the image is not merely visual content, but evidence whose value depends on whether the depicted scene can be trusted, whether relevant details have been altered, and whether the image can be situated within a reliable chain of provenance. This paper addresses that problem as one of evidentiary image authentication, where the task is to evaluate whether photographic evidence is authentic, manipulated, or synthetic under conditions that resemble legal and institutional use.

Photographs have never been self-authenticating records of reality, since courts have long required them to be connected to testimony, context, and procedure before they can function as evidence \cite{grossman_judicial_2025}. Historical accounts of photographic admissibility show that courts have repeatedly had to reconcile the apparent objectivity of photographs with the possibility of staging, selection, distortion, and later digital alteration \cite{mnookin_image_1998}. The current challenge is therefore not that visual evidence has suddenly become uncertain, but that the practical conditions under which courts have managed that uncertainty have changed. Evidence that once required specialized skill to fabricate can now be altered or produced through widely available generative systems.

Fabricated and misleading evidence is not new, yet generative AI changes its cost, accessibility, and plausibility in ways that are directly relevant to evidentiary workflows. Earlier forms of photographic manipulation often required specialized tools, technical expertise, or substantial effort, whereas contemporary image-generation systems allow non-experts to produce plausible visual alterations through ordinary prompt-based interfaces. A litigant, witness, or third party can change scene conditions, insert or remove salient objects, or generate a plausible image from scratch without the expertise that previously constrained visual forgery. The same technological shift also creates a second evidentiary risk, since genuine evidence may be dismissed as fabricated precisely because fabrication itself has become credible. This is the dynamic Chesney and Citron \cite{chesneyDeepFakesLooming2018} describe as the liar’s dividend, and it places pressure on courts and related institutions not only to detect false evidence, but also to preserve confidence in authentic evidence.

Recent legal developments show that this concern has moved beyond speculation, as courts and judicial organizations are now confronting acknowledged and suspected AI-generated material, including fabricated legal authorities, synthetic media, and allegedly manipulated exhibits. In Mata v. Avianca \cite{ryan_practical_2023}, generative AI produced plausible but non-existent legal authorities that were submitted in litigation, illustrating how synthetic material can enter formal legal processes when verification fails. More directly for visual evidence, judicial and legal scholarship has begun to examine how courts should handle acknowledged and unacknowledged AI-generated evidence, while court-focused organizations have warned that synthetic evidence may affect public trust in judicial proceedings \cite{heaton_ai-generated_2026}. These developments make evidentiary image authentication a timely problem for legal institutions and for computational researchers working on information integrity, provenance, verification, and trustworthy decision support.

The evidentiary risk is not limited to fully synthetic images, because in many legal settings the more consequential manipulation is a localized or contextual edit to an otherwise ordinary photograph. A person may be removed from a surveillance frame, a road sign or traffic signal may be changed in a dashcam still, or visible damage may be added to or removed from a phone photograph. Such edits can alter what an image appears to prove while preserving most of its visual content, which makes them difficult for lay observers \cite{bray_testing_2023,roca_how_2025} and also difficult for detectors trained mainly on whole-image synthesis or face-centered deepfakes. For evidentiary authentication, the relevant question is therefore not simply whether an image was generated by AI, but whether verification systems can identify the forms of manipulation that change the informational and evidentiary meaning of a visual item.

Existing datasets do not adequately support this domain. Face-deepfake corpora have enabled important progress in synthetic-media detection, but they primarily address identity, expression, and facial manipulation rather than scene-level photographic evidence. General image-tampering datasets capture classical operations such as splicing, copy-move, and removal, but many predate prompt-driven consumer generation and do not reflect the artifacts, workflows, or editing patterns now available to non-experts. Other public resources provide authentic surveillance, driving, or consumer imagery, yet they do not provide controlled AI-generated manipulations, prompt metadata, generator metadata, or benchmark-ready labels. What is missing is a public resource for evaluating authentication systems on evidentiary image families, realistic manipulation types, consumer-interface generation, structured provenance, and detector behavior under conditions relevant to institutional use.

We introduce the CIFAR Synthetic Evidence Corpus for Detecting AI-Manipulated Images, a dataset and benchmark for evidentiary image authentication. The corpus contains 1,505 photographic items, including 720 authentic controls and 785 manipulated or fabricated images. It spans three families of visual evidence, namely surveillance imagery, dashcam stills, and consumer photographs. Manipulations are organized into three tiers covering scene-condition changes, localized element edits, and complete fabrication. The manipulated images are produced using contemporary generative systems from multiple organizations, and each item is paired with metadata describing its source provenance, evidentiary family, manipulation tier, generator, prompt template, and scene attributes. 

This paper makes four contributions. First, it releases a benchmark corpus of authentic and AI-manipulated photographic evidence across surveillance, dashcam, and consumer-photo imagery. Second, it introduces a manipulation taxonomy grounded in evidentiary practice, with particular attention to localized edits that can alter the meaning of a visual exhibit while leaving most of the scene intact. Third, it provides a generation pipeline based on a realistic non-expert consumer-tool threat model. Fourth, it reports a zero-shot benchmark of off-the-shelf detectors, showing that current systems exhibit error profiles that remain problematic for evidentiary use. The corpus is available at \url{https://github.com/UofT-CIFAR/Synthetic-Evidence-Image-Corpus}, DOI: \href{https://doi.org/10.5281/zenodo.22015105}{10.5281/zenodo.22015105}.

\section{Distinction from Existing Resources}
\label{sec:RelatedWork}
Our corpus complements rather than replaces existing image-forensics benchmarks by targeting the authentication of photographic evidence under contemporary AI-enabled manipulation. Face-manipulation corpora such as FaceForensics++, Celeb-DF, DFDC, ForgeryNet, and OpenForensics \cite{rosslerFaceForensicsLearningDetect2019,liCelebDFLargescaleChallenging2020,dolhansky_deepfake_2020,he_forgerynet_2021,le_openforensics_2021} primarily treat the face as the object of authentication, whereas our corpus treats the photograph as a visual claim about an event, location, object, or condition. General tampering benchmarks such as CASIA and the NIST Media Forensics Challenge datasets \cite{dong_casia_2013,guan_mfc_2019} provide important coverage of splicing, copy-move, removal, and related operations, but largely reflect editing paradigms that predate consumer generative systems. Our corpus instead emphasizes prompt-driven generation and editing, including localized changes, scene-level modifications, and complete fabrication, capturing a contemporary non-expert threat model \cite{corviDetectionSyntheticImages2022,chandraDeepfakeEval2024MultiModalIntheWild2025}. Similarly, authentic-image resources such as VIRAT, UCF-Crime, BDD100K, and CASIA \cite{oh_large-scale_2011,sultani_real-world_2019,yu_bdd100k_2020,dong_casia_2013} provide valuable source imagery but do not pair it with controlled generative edits, generation metadata, and manipulation labels designed for authentication analysis.

Our corpus extends this ecosystem by organizing manipulated photographic evidence according to evidentiary family, manipulation tier and subtype, source dataset, generator, prompt, and scene attributes. This structure supports analysis beyond aggregate detection accuracy, allowing researchers to evaluate which evidentiary changes are detectable, which are missed, and which authentic images generate false positives. This focus addresses a gap identified by legal scholarship, which has increasingly emphasized the need for reliable authentication of AI-generated evidence and stronger mechanisms for addressing synthetic-media risks in litigation \cite{chesneyDeepFakesLooming2018,delfinoDeepfakesTrialCall2023,dalalDeepfakesCourtHow2025,grimm_artificial_2021,grossman_judicial_2025}. While that literature establishes the institutional problem, it does not provide the empirical infrastructure needed to measure authentication performance on evidentiary photographs. Our corpus addresses this missing empirical layer without claiming that detection alone can resolve the legal problem; instead, it provides a benchmark for studying the reliability, failure modes, and limits of authentication systems in a setting aligned with evidentiary practice.

\section{Resource Design and Construction}
\label{sec:Resource}
\begin{figure}[ht]
\centering
\setlength{\tabcolsep}{1pt}
\renewcommand{\arraystretch}{0.9}
\begin{tabular}{@{}c@{\hspace{2pt}}c@{}}
  \includegraphics[width=4.2cm]{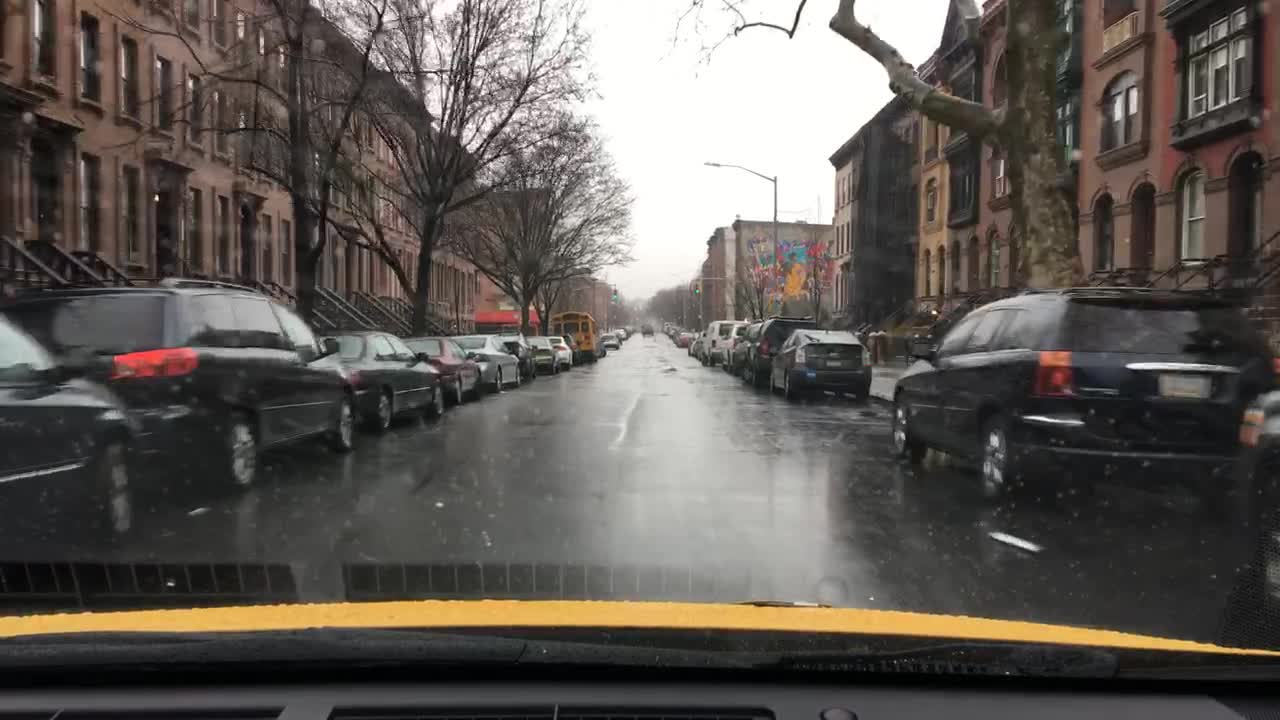} &
  \includegraphics[width=4.2cm]{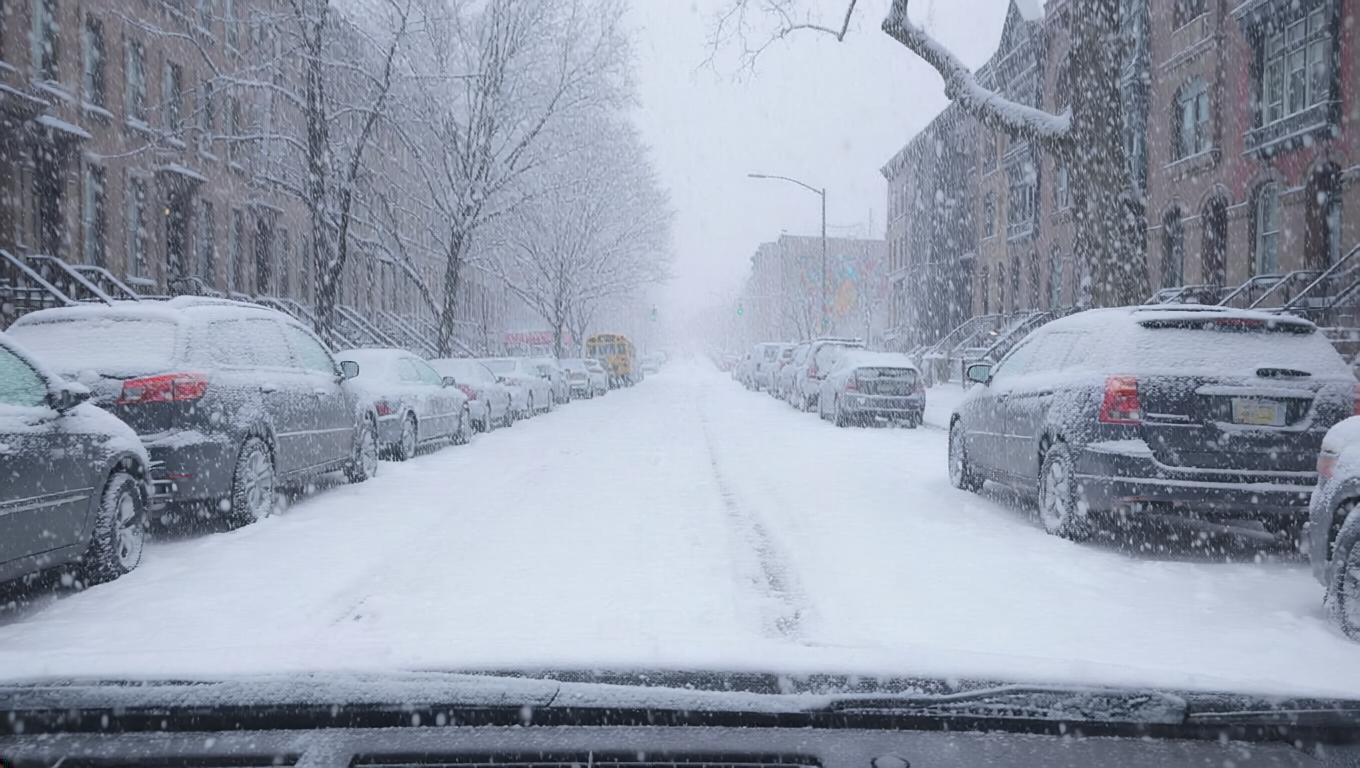} \\
  {\footnotesize (a) Authentic source} & {\footnotesize (b) T1: weather edit} \\
\end{tabular}
\vspace{1pt}
\begin{tabular}{@{}c@{\hspace{2pt}}c@{}}
  \includegraphics[width=4.2cm]{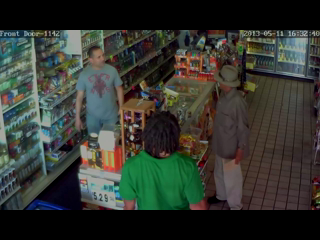} &
  \includegraphics[width=4.2cm]{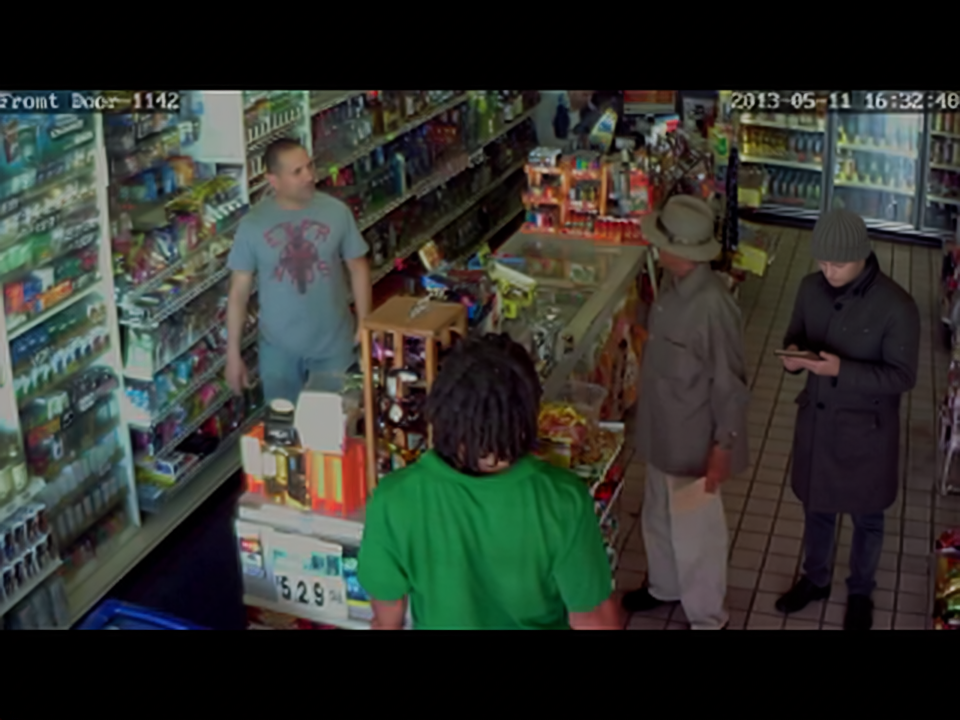} \\
  {\footnotesize (c) Authentic source} & {\footnotesize (d) T2: person added} \\
\end{tabular}
\caption{Example manipulations. \emph{Top:} Tier-1 scene-condition edit adds snow to a dashcam frame. \emph{Bottom:} Tier-2 localized edit inserts a person into a surveillance frame.}
\Description{Four image exemplars from the dataset. The top two images depict before and after of Tier-1 scene-condition edit adds snow to a dashcam frame. The bottom two images depict before and after of Tier-2 localized edit inserts a person into a surveillance frame.}
\label{fig:examples}
\end{figure}
Our proposed corpus is designed for evidentiary image authentication rather than general synthetic-image detection. This setting requires a benchmark that represents the kinds of photographs submitted in legal and institutional contexts, includes manipulations that can change the evidentiary meaning of an image, reflects the consumer-tool threat model through which non-experts can now produce plausible edits, and provides metadata that supports controlled analysis beyond aggregate binary classification.


We operationalize these requirements along four axes. \textbf{First}, the corpus covers three evidentiary families: surveillance, dashcam, and consumer photographs. 
\textbf{Second}, it distinguishes three manipulation tiers: scene-condition
changes (T1), localized element edits (T2), and complete fabrication
(T3); Table~\ref{tab:family_tier} reports the full subtype breakdown by
family. Figure~\ref{fig:examples} shows representative T1 and T2
manipulations.
\textbf{Third}, we use consumer-accessible generative systems from multiple organizations to reduce dependence on artifacts from a single provider. \textbf{Fourth}, each item includes metadata describing its source, evidentiary family, manipulation tier and subtype, generator, prompt, and scene attributes.

For constructing the corpus, we first defined a fixed set of prompt templates, each tied to a family, tier, and subtype. The templates were written to model a non-expert using a consumer image generation interface. Each prompt requests a single scoped change, asks the model to preserve all other scene content, and aims to produce a plausible evidentiary image rather than an arbitrary edit. The final prompt set contains 27 templates spanning the three families and three tiers.

We constructed the source pool from public datasets matched to the three evidentiary families: VIRAT and UCF-Crime for surveillance, BDD100K for dashcam imagery, and CASIA~v2.0 for consumer photographs \cite{oh_large-scale_2011,sultani_real-world_2019,yu_bdd100k_2020,dong_casia_2013}. Tier~1 and Tier~2 manipulations use authentic source frames, while Tier~3 images are generated directly from prompts. Candidate frames were filtered using automatically extracted scene attributes to satisfy predefined quotas, and no source frame is reused.
\begin{table*}[t]
\centering
\small
\setlength{\tabcolsep}{5pt}
\caption{Manipulated image corpus by evidentiary family, source
dataset, and manipulation tier.}
\label{tab:family_tier}
\begin{tabular}{@{}ll ll ll r r@{}}
\toprule
\textbf{Family} & \textbf{Source dataset(s)}
 & \multicolumn{2}{c}{\textbf{T1 (scene conditions)}}
 & \multicolumn{2}{c}{\textbf{T2 (element edit)}}
 & \textbf{T3} & \textbf{Total} \\
\midrule

\multirow{4}{*}{\parbox{2.0cm}{Surveillance\\}} & \multirow{4}{*}{\parbox{2.5cm}{VIRAT, UCF-Crime}} & Time of day & 30 & Swap object & 28 & \multirow{4}{*}{45} & \multirow{4}{*}{\textbf{264}} \\
 &  & Weather & 28 & Remove person & 26 &  &  \\
 &  & Lighting & 29 & Remove signage & 25 &  &  \\
 &  & Crowd density & 30 & Add person & 23 &  &  \\
\cmidrule(lr){1-8}
\multirow{4}{*}{\parbox{2.0cm}{Dashcam}} & \multirow{4}{*}{\parbox{2.5cm}{BDD100K}} & Time of day & 29 & Remove sign & 30 & \multirow{4}{*}{45} & \multirow{4}{*}{\textbf{255}} \\
 &  & Weather & 27 & Traffic-light color & 27 &  &  \\
 &  & Lighting & 29 & Remove vehicle & 27 &  &  \\
 &  & Crowd density & 27 & Add hazard & 14 &  &  \\
\cmidrule(lr){1-8}
\multirow{4}{*}{\parbox{2.0cm}{Consumer\\photos}} & \multirow{4}{*}{\parbox{2.5cm}{CASIA}} & Time of day & 30 & Add object & 29 & \multirow{4}{*}{45} & \multirow{4}{*}{\textbf{266}} \\
 &  & Weather & 30 & Remove damage & 28 &  &  \\
 &  & Lighting & 30 & Swap object & 25 &  &  \\
 &  & Crowd density & 30 & Add damage & 19 &  &  \\
\cmidrule(lr){1-8}
\textbf{Total} & & & \textbf{349} & & \textbf{301} & \textbf{135} & \textbf{785} \\
\bottomrule
\end{tabular}
\end{table*}
Manipulated items were generated by applying the appropriate prompt template to the assigned source frame for Tier~1 and Tier~2, or from the prompt alone for Tier~3. Items were screened during generation and regenerated where the requested edit failed to appear or the scene was visibly degraded; this was not conducted as a formal annotation exercise, so we do not report inter-annotator agreement. The main corpus uses GPT Image~2, Gemini~3.1 with Nano Banana~2, and Firefly~5. GPT Image~2 and Gemini~3.1 were accessed through APIs, while Firefly~5 was accessed through its consumer interface. The corpus also includes a smaller \emph{Hero} set produced through iterative human-in-the-loop refinement in Photoshop and ChatGPT. The main variants model ordinary single-pass manipulation, while the Hero set models a higher-effort adversary willing to manually refine an image.

\section{Dataset Content and Benchmarking}
\label{sec:DatasetBenchmark}
\begin{table}[t]
\centering
\small
\setlength{\tabcolsep}{4pt}
\caption{Performance of off-the-shelf detectors.}
\label{tab:benchmark}
\begin{tabular}{lrrrrrr}
\toprule
Detector & Acc. & Prec. & Rec. & FPR & ROC-AUC & PR-AUC \\
\midrule
\textsc{d3}            & 0.616 & 0.696 & 0.469 & 0.224 & 0.674 & 0.684 \\
\textsc{univfd}        & 0.573 & 0.686 & 0.336 & 0.168 & 0.643 & 0.633 \\
\textsc{drct}          & 0.535 & 0.647 & 0.238 & 0.141 & 0.621 & 0.642 \\
\textsc{cnnspot}       & 0.478 & 0.500 & 0.003 & 0.003 & 0.639 & 0.587 \\
\bottomrule
\end{tabular}
\end{table}
To establish reference points for future work, we evaluate publicly available image-manipulation detectors on the corpus in a zero-shot setting. Each detector scores every image once using its released weights and default decision threshold, with no fine-tuning, and we report both threshold-free metrics (ROC-AUC, PR-AUC), which rank detectors independently of any operating point, and threshold-dependent metrics (accuracy, precision, recall, FPR) at each detector's shipped default. The evaluation uses the DetectZoo framework~\cite{ebrahimi_detectzoo_2026} and treats the task as binary authentication over all 1{,}505 items. The detectors span  released synthetic-image detectors: a diffusion-reconstruction method (\textsc{d3}), a CLIP-feature classifier (\textsc{univfd}), a diffusion-reconstruction contrastive detector (\textsc{drct}), and a CNN-based GAN-artifact detector (\textsc{cnnspot}).

Across all systems, performance is low: the best detector, \textsc{d3}, reaches only 0.674 ROC-AUC, and the rest sit near chance (Table~\ref{tab:benchmark}). This is the central benchmarking result---off-the-shelf detectors, trained predominantly on fully synthesized GAN or diffusion imagery, transfer poorly to the localized edits and consumer-tool manipulations that characterize evidentiary content. Default operating points are also poorly aligned with evidentiary use: at its shipped threshold \textsc{d3} catches fewer than half of all manipulations (47\% recall) yet wrongly flags 22\% of authentic images as manipulated, and falsely impugning genuine evidence is the more damaging error in court. A benchmark for evidentiary authentication must therefore report this error structure, not only aggregate accuracy.

A breakdown of the strongest detector, \textsc{d3}, illustrates the diagnostic value of the corpus design. Performance is uneven in ways that track evidentiary structure rather than overall difficulty. By family, dashcam imagery is markedly harder than surveillance or consumer photographs (AUC 0.55 versus 0.69 and 0.78). By generator, detectability varies sharply: items from one consumer model are far less detectable than another's (per-generator recall ranging from 0.46 to 0.71), confirming that the corpus exposes cross-generator generalization gaps. The high-effort \emph{Hero} set is the most revealing case: manually refined Photoshop edits receive extremely low manipulation scores, the lowest scoring at 0.04, and slip past the detector entirely, direct evidence that a motivated, hands-on adversary can defeat current systems. These findings position the corpus as a diagnostic benchmark for studying not just whether authentication systems succeed, but where they fail and which failure modes matter for evidentiary use.

\section{Ethical Considerations}
The corpus derives from publicly released research datasets and is redistributed in accordance with source licenses. No imagery was collected from human participants and no material originates from real legal proceedings. As the work involves secondary use of data with no participant interaction, it did not require research ethics board review. 
We draw only on the normal-activity subset of UCF-Crime; no individual appears in connection to a crime and no manipulation was designed to place an identifiable person at a real location or to attribute conduct to them. 
Manipulations were produced using capabilities already available through consumer interfaces so we judge the uplift from release to be small relative to the benefit to detection research.

\section{Concluding Remarks}
This paper introduces the CIFAR Synthetic Evidence Corpus for evidentiary image authentication. The corpus enables diagnostic evaluation of contemporary AI manipulations across diverse photographic evidence, while our baselines show substantial weaknesses in current detectors, particularly for localized edits and authentic-image false positives. We release the corpus and benchmark infrastructure to support future research on reliable visual evidence authentication.

\begin{acks}
This work was supported by the \emph{Safeguarding Courts from Synthetic AI Content} Solution Network, funded through the Canadian AI Safety Institute (CAISI) Research Program at CIFAR.
\end{acks}
\section*{GenAI Usage Disclosure}
The generative models central to this work (GPT Image~2 (OpenAI), Gemini~3.1 with Nano Banana~2 (Google), and Firefly~5 (Adobe)) were used to construct the manipulated portion of the corpus, as described in Section~\ref{sec:Resource}. This use is the subject of the resource itself rather than an authoring aid, and all generated artifacts are released with the provenance metadata documented above.

In preparing the manuscript, the authors used Claude to assist with improving the clarity of author-written prose and generating draft LaTeX for tables. All AI-assisted text was reviewed and edited by the authors, who take full responsibility for the content of the paper. No AI system is listed as an author.
\balance
\bibliographystyle{ACM-Reference-Format}
\bibliography{references}

@inproceedings{guan_mfc_2019,
	address = {Waikola, HI},
	title = {{MFC} {Datasets}: {Large}-{Scale} {Benchmark} {Datasets} for {Media} {Forensic} {Challenge} {Evaluation}},
	shorttitle = {{MFC} {Datasets}},
	url = {https://www.nist.gov/publications/mfc-datasets-large-scale-benchmark-datasets-media-forensic-challenge-evaluation},
	language = {en},
	urldate = {2026-02-26},
	booktitle = {{IEEE} {Winter} {Conference} on {Applications} of {Computer} {Vision} ({WACV} 2019)},
	publisher = {IEEE},
	author = {Guan, Haiying and Kozak, Mark and Robertson, Eric and Lee, Yooyoung and Yates, Amy and Delgado, Andrew and Zhou, Daniel F. and Kheyrkhah, Timothée N. and Smith, Jeff and Fiscus, Jonathan G.},
	month = jan,
	year = {2019},
	note = {Last Modified: 2019-12-31T06:12-05:00},
}

@article{ryan_practical_2023,
	address = {United States of America},
	title = {Practical {Lessons} from the {Attorney} {AI} {Missteps} in {Mata} v. {Avianca}},
	url = {https://www.acc.com/resource-library/practical-lessons-attorney-ai-missteps-mata-v-avianca},
	language = {en},
	urldate = {2026-06-06},
	journal = {Association of Corporate Counsel},
	author = {Ryan, William A. and Garrett, Allen},
	month = aug,
	year = {2023},
}

@misc{chandraDeepfakeEval2024MultiModalIntheWild2025,
	title = {Deepfake-{Eval}-2024: {A} {Multi}-{Modal} {In}-the-{Wild} {Benchmark} of {Deepfakes} {Circulated} in 2024},
	shorttitle = {Deepfake-{Eval}-2024},
	url = {http://arxiv.org/abs/2503.02857},
	doi = {10.48550/arXiv.2503.02857},
	urldate = {2026-02-20},
	publisher = {arXiv},
	author = {Chandra, Nuria Alina and Murtfeldt, Ryan and Qiu, Lin and Karmakar, Arnab and Lee, Hannah and Tanumihardja, Emmanuel and Farhat, Kevin and Caffee, Ben and Paik, Sejin and Lee, Changyeon and Choi, Jongwook and Kim, Aerin and Etzioni, Oren},
	month = mar,
	year = {2025},
	note = {arXiv:2503.02857 [cs]
version: 1},
}

@article{heaton_ai-generated_2026,
	title = {{AI}-generated evidence is a threat to public trust in the courts {\textbar} {National} {Center} for {State} {Courts}},
	url = {https://www.ncsc.org/resources-courts/ai-generated-evidence-threat-public-trust-courts},
	language = {en},
	urldate = {2026-06-06},
	journal = {National Center for State Courts},
	author = {Heaton, Connor and Cleary, Shay and Navin, Michael},
	month = feb,
	year = {2026},
}

@inproceedings{oh_large-scale_2011,
	address = {Colorado Springs, Colorado},
	title = {A large-scale benchmark dataset for event recognition in surveillance video},
	issn = {1063-6919},
	url = {https://ieeexplore.ieee.org/document/5995586},
	doi = {10.1109/CVPR.2011.5995586},
	urldate = {2026-06-06},
	booktitle = {{CVPR} 2011},
	publisher = {CVPR 2011},
	author = {Oh, Sangmin and Hoogs, Anthony and Perera, Amitha and Cuntoor, Naresh and Chen, Chia-Chih and Lee, Jong Taek and Mukherjee, Saurajit and Aggarwal, J. K. and Lee, Hyungtae and Davis, Larry and Swears, Eran and Wang, Xioyang and Ji, Qiang and Reddy, Kishore and Shah, Mubarak and Vondrick, Carl and Pirsiavash, Hamed and Ramanan, Deva and Yuen, Jenny and Torralba, Antonio and Song, Bi and Fong, Anesco and Roy-Chowdhury, Amit and Desai, Mita},
	month = jun,
	year = {2011},
	pages = {3153--3160},
}

@inproceedings{le_openforensics_2021,
	address = {Montreal, QC},
	title = {{OpenForensics}: {Large}-{Scale} {Challenging} {Dataset} {For} {Multi}-{Face} {Forgery} {Detection} {And} {Segmentation} {In}-{The}-{Wild}},
	isbn = {978-1-6654-2812-5},
	shorttitle = {{OpenForensics}},
	url = {https://www.computer.org/csdl/proceedings-article/iccv/2021/281200k0097/1BmEUqCClbi},
	doi = {10.1109/ICCV48922.2021.00996},
	language = {English},
	urldate = {2026-06-06},
	booktitle = {2021 {IEEE}/{CVF} {International} {Conference} on {Computer} {Vision} ({ICCV})},
	publisher = {IEEE Computer Society},
	author = {Le, Trung-Nghia and Nguyen, Huy H. and Yamagishi, Junichi and Echizen, Isao},
	month = oct,
	year = {2021},
	pages = {10097--10107},
}

@misc{ebrahimi_detectzoo_2026,
	title = {{DetectZoo}: {A} {Unified} {Toolkit} for {AI}-{Generated} {Content} {Detection} {Across} {Text}, {Audio}, and {Image} {Modalities}},
	shorttitle = {{DetectZoo}},
	url = {http://arxiv.org/abs/2606.04205},
	doi = {10.48550/arXiv.2606.04205},
	urldate = {2026-06-06},
	publisher = {arXiv},
	author = {Ebrahimi, Sajad and Jamali, Nima and Shirsalimian, Bardia and McConvey, Kelly and Zhang, Wentao and Mahdavimoghaddam, Jalehsadat and Taranukhin, Maksym and Grossman, Maura and Shwartz, Vered and Deng, Yuntian and Bagheri, Ebrahim},
	month = jun,
	year = {2026},
	note = {arXiv:2606.04205 [cs.MM]},
}

@misc{mnookin_image_1998,
	address = {Rochester, NY},
	type = {{SSRN} {Scholarly} {Paper}},
	title = {The {Image} of {Truth}: {Photographic} {Evidence} and the {Power} of {Analogy}},
	shorttitle = {The {Image} of {Truth}},
	url = {https://papers.ssrn.com/abstract=135311},
	language = {en},
	urldate = {2026-06-06},
	publisher = {Social Science Research Network},
	author = {Mnookin, Jennifer},
	month = feb,
	year = {1998},
}

@article{grossman_judicial_2025,
	title = {Judicial {Approaches} to {Acknowledged} and {Unacknowledged} {AI}-{Generated} {Evidence}},
	volume = {26},
	copyright = {Copyright (c) 2025 Maura Grossman, Paul Grimm},
	issn = {1938-0976},
	url = {https://journals.library.columbia.edu/index.php/stlr/article/view/13890},
	doi = {10.52214/stlr.v26i2.13890},
	language = {en},
	number = {2},
	urldate = {2026-04-23},
	journal = {Science and Technology Law Review},
	author = {Grossman, Maura and Grimm, Paul},
	month = may,
	year = {2025},
}

@article{grimm_artificial_2021,
	title = {Artificial {Intelligence} as {Evidence}},
	volume = {19},
	issn = {1549-8271},
	url = {https://scholarlycommons.law.northwestern.edu/njtip/vol19/iss1/2},
	number = {1},
	journal = {Northwestern Journal of Technology and Intellectual Property},
	author = {Grimm, Paul and Grossman, Maura and Cormack, Gordon},
	month = dec,
	year = {2021},
	pages = {9},
}

@misc{yu_bdd100k_2020,
	title = {{BDD100K}: {A} {Diverse} {Driving} {Dataset} for {Heterogeneous} {Multitask} {Learning}},
	shorttitle = {{BDD100K}},
	url = {http://arxiv.org/abs/1805.04687},
	doi = {10.48550/arXiv.1805.04687},
	urldate = {2026-02-26},
	publisher = {arXiv},
	author = {Yu, Fisher and Chen, Haofeng and Wang, Xin and Xian, Wenqi and Chen, Yingying and Liu, Fangchen and Madhavan, Vashisht and Darrell, Trevor},
	month = apr,
	year = {2020},
	note = {arXiv:1805.04687 [cs]},
}

@misc{sultani_real-world_2019,
	title = {Real-world {Anomaly} {Detection} in {Surveillance} {Videos}},
	url = {http://arxiv.org/abs/1801.04264},
	doi = {10.48550/arXiv.1801.04264},
	urldate = {2026-02-26},
	publisher = {arXiv},
	author = {Sultani, Waqas and Chen, Chen and Shah, Mubarak},
	month = feb,
	year = {2019},
	note = {arXiv:1801.04264 [cs]},
}

@inproceedings{dong_casia_2013,
	title = {{CASIA} {Image} {Tampering} {Detection} {Evaluation} {Database}},
	url = {https://ieeexplore.ieee.org/document/6625374},
	doi = {10.1109/ChinaSIP.2013.6625374},
	urldate = {2026-02-26},
	booktitle = {2013 {IEEE} {China} {Summit} and {International} {Conference} on {Signal} and {Information} {Processing}},
	author = {Dong, Jing and Wang, Wei and Tan, Tieniu},
	month = jul,
	year = {2013},
	pages = {422--426},
}

@misc{dolhansky_deepfake_2020,
	title = {The {DeepFake} {Detection} {Challenge} ({DFDC}) {Dataset}},
	url = {http://arxiv.org/abs/2006.07397},
	doi = {10.48550/arXiv.2006.07397},
	urldate = {2026-02-26},
	publisher = {arXiv},
	author = {Dolhansky, Brian and Bitton, Joanna and Pflaum, Ben and Lu, Jikuo and Howes, Russ and Wang, Menglin and Ferrer, Cristian Canton},
	month = oct,
	year = {2020},
	note = {arXiv:2006.07397 [cs]},
}

@misc{he_forgerynet_2021,
	title = {{ForgeryNet}: {A} {Versatile} {Benchmark} for {Comprehensive} {Forgery} {Analysis}},
	shorttitle = {{ForgeryNet}},
	url = {http://arxiv.org/abs/2103.05630},
	doi = {10.48550/arXiv.2103.05630},
	urldate = {2026-02-26},
	publisher = {arXiv},
	author = {He, Yinan and Gan, Bei and Chen, Siyu and Zhou, Yichun and Yin, Guojun and Song, Luchuan and Sheng, Lu and Shao, Jing and Liu, Ziwei},
	month = jul,
	year = {2021},
	note = {arXiv:2103.05630 [cs]},
}

@misc{liCelebDFLargescaleChallenging2020,
	title = {Celeb-{DF}: {A} {Large}-scale {Challenging} {Dataset} for {DeepFake} {Forensics}},
	shorttitle = {Celeb-{DF}},
	url = {http://arxiv.org/abs/1909.12962},
	doi = {10.48550/arXiv.1909.12962},
	urldate = {2026-02-20},
	publisher = {arXiv},
	author = {Li, Yuezun and Yang, Xin and Sun, Pu and Qi, Honggang and Lyu, Siwei},
	month = mar,
	year = {2020},
	note = {arXiv:1909.12962 [cs]},
}

@misc{corviDetectionSyntheticImages2022,
	title = {On the detection of synthetic images generated by diffusion models},
	url = {http://arxiv.org/abs/2211.00680},
	doi = {10.48550/arXiv.2211.00680},
	urldate = {2026-02-20},
	publisher = {arXiv},
	author = {Corvi, Riccardo and Cozzolino, Davide and Zingarini, Giada and Poggi, Giovanni and Nagano, Koki and Verdoliva, Luisa},
	month = nov,
	year = {2022},
	note = {arXiv:2211.00680 [cs]},
}

@misc{chesneyDeepFakesLooming2018,
	address = {Rochester, NY},
	type = {{SSRN} {Scholarly} {Paper}},
	title = {Deep {Fakes}: {A} {Looming} {Challenge} for {Privacy}, {Democracy}, and {National} {Security}},
	shorttitle = {Deep {Fakes}},
	url = {https://papers.ssrn.com/abstract=3213954},
	doi = {10.2139/ssrn.3213954},
	language = {en},
	urldate = {2026-02-20},
	publisher = {Social Science Research Network},
	author = {Chesney, Robert and Citron, Danielle Keats},
	month = jul,
	year = {2018},
}

@misc{rosslerFaceForensicsLearningDetect2019,
	title = {{FaceForensics}++: {Learning} to {Detect} {Manipulated} {Facial} {Images}},
	shorttitle = {{FaceForensics}++},
	url = {http://arxiv.org/abs/1901.08971},
	doi = {10.48550/arXiv.1901.08971},
	urldate = {2026-02-20},
	publisher = {arXiv},
	author = {Rössler, Andreas and Cozzolino, Davide and Verdoliva, Luisa and Riess, Christian and Thies, Justus and Nießner, Matthias},
	month = aug,
	year = {2019},
	note = {arXiv:1901.08971 [cs]},
}

@article{dalalDeepfakesCourtHow2025,
	title = {Deepfakes in {Court}: {How} {Judges} {Can} {Proactively} {Manage} {Alleged} {AI}-{Generated} {Material} in {National} {Security} {Cases}},
	volume = {2024},
	issn = {0892-5593},
	shorttitle = {Deepfakes in {Court}},
	url = {https://chicagounbound.uchicago.edu/uclf/vol2024/iss1/3},
	number = {1},
	journal = {University of Chicago Legal Forum},
	author = {Dalal, Abhishek and Gao, Chongyang and Grimm, Hon Paul and Grossman, Maura and Jr, Daniel Linna and Pulice, Chiara and Subrahmanian, V. S. and Tunheim, Hon John},
	month = jan,
	year = {2025},
}

@article{delfinoDeepfakesTrialCall2023,
	title = {Deepfakes on {Trial}: {A} {Call} {To} {Expand} the {Trial} {Judge}’s {Gatekeeping} {Role} {To} {Protect} {Legal} {Proceedings} from {Technological} {Fakery}},
	volume = {74},
	issn = {0017-8322{\textless}br /{\textgreater}© Copyright University of California, College of the Law San Francisco},
	shorttitle = {Deepfakes on {Trial}},
	url = {https://repository.uclawsf.edu/hastings_law_journal/vol74/iss2/3},
	number = {2},
	journal = {UC Law Journal},
	author = {Delfino, Rebecca},
	month = feb,
	year = {2023},
	pages = {293},
}

@article{bray_testing_2023,
	title = {Testing human ability to detect ‘deepfake’ images of human faces},
	volume = {9},
	issn = {2057-2085},
	url = {https://doi.org/10.1093/cybsec/tyad011},
	doi = {10.1093/cybsec/tyad011},
	number = {1},
	urldate = {2026-02-13},
	journal = {Journal of Cybersecurity},
	author = {Bray, Sergi D and Johnson, Shane D and Kleinberg, Bennett},
	month = jan,
	year = {2023},
	pages = {tyad011},
}

@misc{roca_how_2025,
	title = {How good are humans at detecting {AI}-generated images? {Learnings} from an experiment},
	shorttitle = {How good are humans at detecting {AI}-generated images?},
	url = {http://arxiv.org/abs/2507.18640},
	doi = {10.48550/arXiv.2507.18640},
	urldate = {2026-02-13},
	publisher = {arXiv},
	author = {Roca, Thomas and Roman, Anthony Cintron and Vega, Jehú Torres and Duarte, Marcelo and Wang, Pengce and White, Kevin and Misra, Amit and Ferres, Juan Lavista},
	month = may,
	year = {2025},
	note = {arXiv:2507.18640 [cs]
version: 1},
}
\end{document}